\documentclass[10pt,twocolumn,letterpaper]{article}

\usepackage[pagenumbers]{cvpr} 

\usepackage[dvipsnames]{xcolor}
\usepackage{soul}     

\usepackage{booktabs}
\usepackage{times}
\usepackage{fancyhdr,graphicx,amsmath,amssymb}
\usepackage{algorithm}
\usepackage{algpseudocode}
\usepackage{amsmath}
\usepackage{gensymb}
\usepackage{dsfont}

\newif\ifshowedits

\newcommand{\addeditor}[3]{%
  \definecolor{#1color}{rgb}{#3}
  \expandafter\newcommand\csname #1\endcsname[1]{%
  \ifshowedits
    {\color{#1color} ##1}%
  \else
    {##1}%
  \fi
  }%
  \expandafter\newcommand\csname #1rmk\endcsname[1]{%
  \ifshowedits
    {\color{#1color} {\bf [#2: ##1]}}
  \fi
  }%
  \expandafter\newcommand\csname #1rpl\endcsname[2]{%
  \ifshowedits
    {{\color{#1color} ##1} \sout{##2}}
  \else
    {##1}
  \fi
  }%
}

\newcommand{\mycomment}[1]{}

\usepackage{soul}
\usepackage{amsmath}
\usepackage{amssymb}
\usepackage{wrapfig}
\usepackage{booktabs}
\usepackage{multirow}
\usepackage{xspace}

\NewDocumentCommand{\rot}{O{45}
O{1em}m}{\makebox[#2][l]{\rotatebox{#1}{#3}}}%

\def\mycfigure#1#2{%
    \begin{figure*}[htb]%
    \centering\includegraphics*[width = \linewidth]{figures/#1}%
    \vspace{-.2cm}%
    \caption{#2}%
    \vspace{-.3cm}%
    \label{fig:#1}%
    \end{figure*}%
}

\newcommand{\refFig}[1]{Fig.~\ref{fig:#1}}

\newcommand{\refAlg}[1]{Alg.~\ref{alg:#1}}

\newcommand{\change}[1]{#1}

\soulregister\ref7
\soulregister\cite7
\soulregister\refFig7
\soulregister\refAlg7
\soulregister\cite7
\soulregister\ref7
\soulregister\pageref7
\soulregister\shortcite7
\soulregister\eg0
\soulregister\ie0
\soulregister\etal0

\DeclareGraphicsExtensions{.png,.jpg,.pdf,.ai,.psd}
\definecolor{colorA}{HTML}{4285f4}
\definecolor{colorB}{HTML}{ea4335}
\definecolor{colorC}{HTML}{fbbc04}
\definecolor{colorD}{HTML}{34a853}
\definecolor{colorE}{HTML}{ff6d01}
\definecolor{colorF}{HTML}{46bdc6}
\definecolor{colorG}{HTML}{000000}
\definecolor{colorH}{HTML}{777777}
\definecolor{colorI}{HTML}{bdd6ff}
\definecolor{colorJ}{HTML}{6a9e6f}

\usepackage{pifont}
\definecolor{cvprblue}{rgb}{0.21,0.49,0.74}
\usepackage[pagebackref,breaklinks,colorlinks,citecolor=cvprblue]{hyperref}

\usepackage{xr}

\def\paperID{111} 
\def\confName{3DV\xspace}
\def\confYear{2026\xspace}

\title{SAMa: Material-Aware 3D Selection and Segmentation}

\author{
    Michael Fischer$^{1,2}$\thanks{Corresponding author. Work done during an internship at Adobe Research. Contact: m.fischer@cs.ucl.ac.uk.}\,\,, 
    Iliyan Georgiev$^{1}$, 
    Thibault Groueix$^{1}$, 
    Vladimir G. Kim$^{1}$,
    \vspace{0.15cm} \\ 
    Tobias Ritschel$^{2}$, 
    Valentin Deschaintre$^{1}$
    \vspace{0.3cm} \\ 
    $^{1}$Adobe Research \quad $^{2}$University College London
}

\usepackage[nolist]{acronym}
\begin{acronym}
\acro{ANN}{approximate nearest neighbour}
\acro{IOU}{intersection over union}
\acro{kNN}{k-nearest neighbour}
\acro{MAE}{masked auto-encoder}
\acro{mIoU}{mean intersection over union}
\acro{NeRF}{Neural Radiance Field}
\acro{NN}{nearest neighbour}
\acro{NVS}{novel view synthesis}
\acro{PBR}{physically-based rendering}
\acro{SAM}{Segment Anything Model}
\acro{ViT}{Vision Transformer}
\end{acronym}

\begin{document}

\definecolor{darkgreen}{RGB}{0,110,0}
\definecolor{darkred}{RGB}{170,0,0}
\def\greencheckmark{\textcolor{darkgreen}{\checkmark}}
\def\redxmark{\textcolor{darkred}{\text{\ding{55}}}}  %

\addeditor{valentin}{VD}{0.7, 0.0, 0.7}
\addeditor{vde}{VD}{0.7, 0.0, 0.7}
\addeditor{thibault}{TG}{0.0, 0.0, 0.8}
\addeditor{vova}{VK}{0.0, 0.5, 0.0}
\addeditor{michael}{MF}{0.1, 0.5, 0.9}
\addeditor{iliyan}{IG}{0.8, 0.0, 0.0}
\addeditor{tobias}{TR}{0.2, 0.8, 0.1}

\showeditsfalse
\showeditstrue


\makeatletter
\g@addto@macro\@maketitle{
    \begin{figure}[H]
        \vspace{-13mm}
        \setlength{\linewidth}{\textwidth}
        \setlength{\hsize}{\textwidth}
        \centering
        \includegraphics{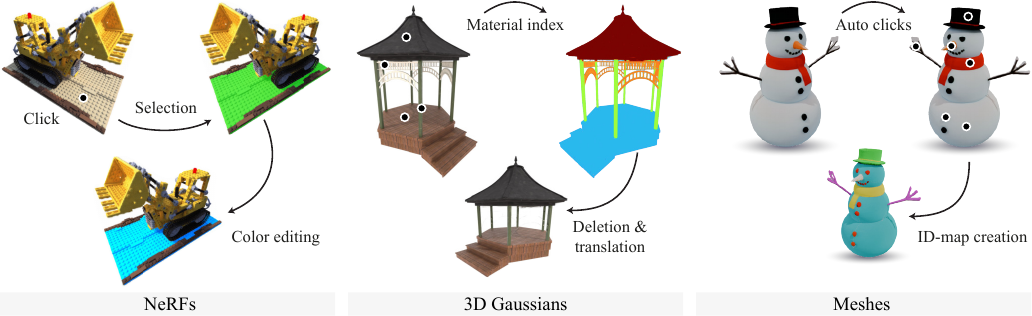}
        \vspace{-6mm}
        \caption{
            3D material selection on three different representations using our SAMa method. Our approach enables several applications; from left to right: color editing on NeRFs, decomposition and editing on Gaussians, automatic material-ID-map creation on meshes.
        }
        \label{fig:Teaser}
    \end{figure}
}
\makeatother

\maketitle

\vspace{-0.5cm}
\begin{abstract}
Decomposing 3D assets into material parts is a common task for artists, yet remains a highly manual process. 
In this work, we introduce \underline{S}elect \underline{A}ny \underline{Ma}terial (SAMa), a material selection approach for in-the-wild objects in arbitrary 3D representations. 
Building on SAM2's video prior, we construct a material-centric video dataset that extends it to the material domain. We propose an efficient way to lift the model's 2D predictions to 3D by projecting each view into an intermediary 3D point cloud using depth.
Nearest-neighbor lookups between any 3D representation and this similarity point cloud allow us to efficiently reconstruct accurate selection masks over objects' surfaces that can be inspected from any view.
Our method is multiview-consistent by design, alleviating the need for costly 
per-asset optimization,
and performs optimization-free selection in seconds. 
SAMa outperforms several strong baselines in selection accuracy and multiview consistency and enables various compelling applications, such as replacing the diffuse-textured materials on a text-to-3D output with PBR materials or selecting and editing materials on NeRFs and 3DGS captures.
Project page: \footnotesize\url{https://mfischer-ucl.github.io/sama/}.
\end{abstract}

\vspace{-1.3cm}
\section{Introduction}
\label{sec:intro}

Understanding the materials around us is an extremely common task for humans, but remains challenging for machines. In this paper we focus on material selection in 3D.

Existing work on material understanding has mostly focused on the 2D image domain, addressing tasks like segmentation \cite{sterbentz2021universal, masubuchi2020deep, upchurch2022dense, bell2015material}, reconstruction \cite{deschaintre2018single, deschaintre2019flexible, shi2020match, hu2022inverse}, generation \cite{vecchio2023controlmat, sartor2023matfusion, vecchio2024stablematerials} or, more recently, material selection \cite{sharma2023materialistic}.  
Semantic segmentation and classification aim to separate areas into different \emph{predetermined} classes, \eg, wood or plastic. This neither accounts for unforeseen materials nor the separation of two materials with different texture properties (\eg, two types of wood).
This work, in contrast, targets \emph{selection}, which aims at finding materials with the same appearance, and thus is more flexible as it can handle any material, regardless of class, and make intra-class distinctions such as two plastics with different appearance.

We follow the material definition of established works \cite{sharma2023materialistic, dana1999reflectance, guerrero2025fine} and consider two materials similar only if they share the same texture and reflectance properties.

Material selection becomes especially relevant in the light of recent generative 3D asset creation and image/text-to-3D workflows. Current methods either provide non-parametric implicit representations (\eg, \acp{NeRF}) or unstructured output (as in triangle soups and baked textures produced by image/text-to-mesh methods~\cite{hong2023lrm, liu2024one, gao2024cat3d, poole2022dreamfusion}),  both of which are challenging to use for artists and downstream tasks. 
Material selection, in this context, has a wide range of downstream applications, \eg, enhancing the X-to-3D workflow with material masks, improving the editability of 3D reconstructions (\eg, through material replacement~\cite{zhang2024mapa, matatlas_Ceylan_2024}), or extracting areas of similar materials as a prior for
inverse rendering~\cite{lensch2003}.

However, most models targeting material-related tasks, including selection, do not trivially extend to the 3D domain, as they are trained on 2D images and therefore have no incentive for producing multiview-consistent predictions~\cite{el2024probing}. Moreover, the 3D domain contains inherent challenges and ambiguities like self- or dis-occlusions and view-dependent effects, and requires accurate propagation of the model predictions into novel, unseen views.
Recent research has therefore developed algorithms to address the problems from multiview-inconsistent predictions that arise when lifting 2D (object) selection to 3D, predominantly via pre-processing noise-consolidation steps such as feature-field distillation \cite{kobayashi2022decomposing} or contrastive (similarity) learning \cite{kim_garfield_2024}, both of which are time-consuming (the former reports a 200k-step, the latter a 20-minute optimization) and must be repeated \emph{per asset}.

In this work, we close this gap between material selection in 2D and 3D by introducing \underline{S}elect \underline{A}ny \underline{Ma}terial (SAMa), an efficient and accurate material selection and segmentation method for 3D assets (see \cref{fig:Teaser}).
Our first core insight is that we can draw parallels between video- and 3D-selection, since in both video and 3D, the selected elements have to be consistent across frames (or views), regardless of object and camera movement or differences in shading and occlusions. 
We thus propose to re-purpose SAM2's recent progress in object selection across video~\cite{ravi2024sam2} for materials. 
We achieve this by fine-tuning parts of the model for (video) material selection on a custom-made video dataset with dense per-pixel, per-frame annotations and show that using videos for fine-tuning is key to achieving high quality in 3D.

Our approach is inherently multiview-consistent thanks to its video training, which lets us avoid costly per-asset optimizations that lift the 2D signal to 3D and allows us resort to a simpler and more efficient strategy instead: for each pixel of each view, we project the 2D similarity information to 3D using the depth and simple inverse camera projections, thus constructing a 3D point cloud aligned with the 3D representation by construction.  
We achieve interactive visualization of our selection results by querying the nearest neighbor between the 3D representation and this similarity point cloud during rendering. 
As a result, our approach supports selection on any 3D representation that can be rendered to an image and queried for depth, such as meshes, Gaussian Splats and NeRFs. 
Additionally, it is fast: the similarly cloud reconstruction can be performed on-the-fly, while the selection visualization from novels views takes less than 10 milliseconds.

In summary, once our model is trained, it enables selection via multi-view renderings of an arbitrary 3D object in less than two seconds from the initial click.


We evaluate our method on meshes, radiance fields and 3D Gaussians, in terms of selection quality and 3D consistency, and show that it improves significantly over existing work and several strong baselines. Finally, we demonstrate multiple applications such as object segmentation into material IDs and NeRF/Gaussian editing.

In summary, we make the following contributions:
\begin{itemize}
    \item Adaptation of a video-object-selection model to material selection on 3D shapes, by training on a novel rendered video dataset which will be released upon acceptance. 
    \item Fast and efficient 3D projections and queries, enabled by cross-frame consistent model predictions.
    \item Multi-modality support, segmentation and editing.
\end{itemize}

Throughout this paper, we will show the user-provided input clicks with respect to which we select materials as black-rimmed circles, and the material similarity to these clicks in false colors, with blue and red indicating low and high, respectively.

\section{Related work}
\label{sec:related_work}

Most related to our work are approaches for material selection on images and approaches that lift a 2D signal defined on renderings into a 3D representation.

\paragraph{Material segmentation datasets.}

Several semantic material datasets with material segmentation annotations exist. 
The Multi-Illumination dataset~\cite{murmann2019dataset}, the Light-Field Material~\cite{wang20164d} and  Flickr Material~\cite{Sharan-JoV-14} datasets respectively contain 1k, 1.2k, and 1k images, segmented with 35, 12 and 10 materials respectively. Of greater size, the OpenSurface~\cite{bell2013opensurfaces},  Material in Context~\cite{bell2015material}, Dense Material Segmentation ~\cite{upchurch2022dense} datasets respectively contain 19k, 437k and 45k images annotated with respectively  37, 23 and 52 types of materials.
The Local Material Database~\cite{schwartz2019recognizing} further annotates 16 kinds of materials on images sources for the previously mentioned datasets. 
These datasets only contain coarse material categories, \eg, two types of metal would have the same ``metal'' label, creating false positives where pixels are marked as sharing the same superclass material but do not share the same appearance.

Materialistic~\cite{sharma2023materialistic} provides a synthetic dataset of 50k HDR images, path-traced from 100 indoor scenes from the Archinteriors collection~\cite{evermotion} and 3k materials.
Complementing this data, \citet{eppel2024learningdataset} extract textures from the Open Images v7 dataset~\cite{openimage} and apply them to random parts of 3D objects from the ShapeNet repository~\cite{shapenet}. The resulting dataset has the advantage of having fine-grained annotations for each material, such as dirt and paint splashes.

Importantly, these datasets~\cite{sharma2023materialistic, eppel2024learningdataset} contain only static renderings, making it challenging to learn multiview consistency. In contrast, our video dataset has dense, fine-grained per-pixel material annotations, enabling the fine-tuning of video selection models.

\paragraph{Material selection.}

Most prior works in material segmentation rely on hand-crafted features~\cite{reyes2006bhattacharyya, belongie1998color, malpica2003multichannel, haralick1973textural, randen1999filtering} or focus on images of flat surfaces~\cite{Lawrence2006, hu2022inverse, Lepage11, Pellacini07, knn_matting}. Recently, \citet{sharma2023materialistic} proposed Materialistic, a model based on DINO-ViT~\cite{caron2021emerging} features, trained to predict the material similarity between a query pixel and all other pixels in a natural image. We find that Materialistic struggles with accurate material selection on 3D objects for two reasons: (1)~it is trained for full-scene photographs, leading to limited selection precision on objects, and (2)~its selections are not sufficiently consistent to be lifted to varying 3D views.
For 3D segmentation, MatSeg3D~\cite{li2024materialseg3d} focuses on per-pixel material classification into a coarse set of predetermined categories (wood, metal, ...), which prevents it from distinguishing between materials within the same category but different appearance (e.g. two different plastics).

Closely related, the \ac{SAM} \cite{sam} uses a ViT trained to predict similarity between pixels. 
As its training data is object-selection specific and not material-aware, SAM requires many separate clicks to perform even moderately well on materials. 
The more recent SAM2~\cite{ravi2024sam2} also targets object selection, but introduces support for temporally coherent predictions across video frames. 
We find that neither SAM, SemanticSAM~\cite{li2024segment, fang2024makeitreal}, nor SAM2 perform well on material selection, except in the special case of an object made of a single material. 
However, once fine-tuned for materials, SAM2's cross-frame selection consistency enables our fast selection lifting to 3D. 

\paragraph{Lifting 2D features to 3D.}

Due to the scarcity of annotated 3D data and the increased computational complexity compared to 2D, many approaches attempt to lift 2D predictions from multiple views to a shared 3D representation. 

The core issue is that the underlying 2D vision models like SAM or DINO are not multiview-consistent. That is, they provide differing predictions for the same 3D point viewed from different positions, making aggregation in 3D challenging. 
Neural Feature Fusion Fields~\cite{neuralfusionfeaturefield} and Feature Field Distillation~\cite{kobayashi2022decomposing} propose to equip NeRFs with an auxiliary feature space, rendered volumetrically to match DINO~\cite{caron2021emerging} or CLIP~\cite{radford2021learning} features. 
Even though the 2D feature maps are not multiview consistent, the shared 3D representation acts as a regularizer and consolidates the quality of the features in rendered novel views \cite{fischer2024nerf}.
This approach has been extended to 3D Gaussian splatting \cite{kerbl20233d} and other image models such as SAM~\cite{Goel_2023_CVPR, lyu2024total, qiu-2024-featuresplatting, Zhou_2024_CVPR, liu2023weakly, lerf2023}. 

Other approaches use contrastive learning to lift segmentation to 3D by pushing closer rendered features of pixels belonging to the same segment, and vice versa~\cite{kim_garfield_2024, ying_omniseg3d_2023, choi2024click, cen2023segment, gu2024egolifter, bhalgat2023contrastive, cen2023saga, liu2024part123, fan2022nerf, ren2022neural}. Our approach differs from this line of work in several ways. First, we lift 2D \emph{material} similarity (rather than object similarity) to 3D. Second, as opposed to previous work, our similarity maps are already multiview consistent thanks to our fine-tuning of a video selection model~\cite{ravi2024sam2}. 
Using this property, we propose a 3D representation-agnostic, lightweight 2D-to-3D lifting approach that does not require any pre-processing.
Contrary to prior work, this allows us to process arbitrary 3D representations (\eg, NeRFs, Gaussians, meshes) and reduces the initial click-to-selection time from 2 hours \cite{neuralfusionfeaturefield} or 20 minutes \cite{kim_garfield_2024} for existing methods to around two seconds.

\mycfigure{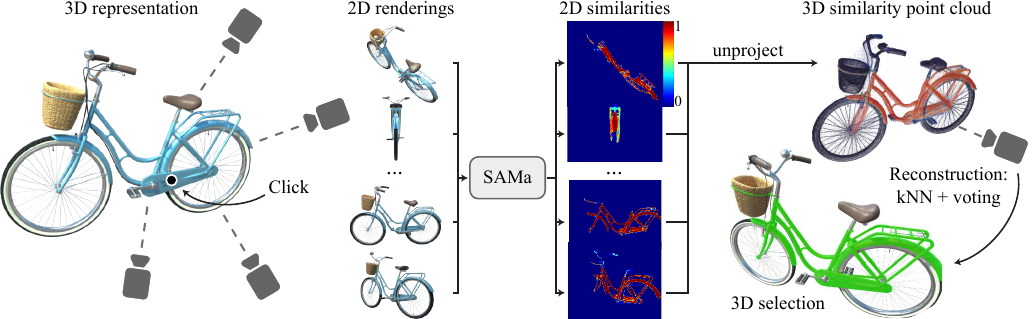}{Overview over our method.
Starting from a 3D asset and a user click, we sample cameras and create a set of renderings covering the object, which we subsequently process with our similarity network SAMa to compute dense per-pixel similarity values. 
We then back-project these values to 3D and store them in a point cloud than can be efficiently queried and interpolated for novel views.}

\section{Method}
\label{sec:method}

Our approach targets \emph{material selection} on 3D representations. Given a 3D asset and a user click, we select all regions of the 3D asset sharing the material appearance of the clicked region. Existing methods focus mostly on selection in 2D images~\cite{sharma2023materialistic}, and their extension to 3D is not trivial due to a lack of selection consistency across views.

Instead of enforcing 3D consistency through per-asset optimization and feature consolidation~\cite{kobayashi2022decomposing, kim_garfield_2024, gu2024egolifter, ying_omniseg3d_2023}, we note that recent video models \cite{ravi2024sam2, valevski2024diffusionmodelsrealtimegame} show good cross-frame consistency.
Since renders of a 3D object from a smooth camera trajectory are not markedly different from a video, 
we propose to adapt SAM2~\cite{ravi2024sam2} to material selection by fine-tuning it, including its memory bank components, on our material-specific video dataset.
Once fine-tuned, the model consumes an image or video and outputs a per-frame floating-point map that encodes the similarity between the clicked pixel's material and all other pixels.

However, fine-tuning is not sufficient for \emph{interactive} selection in 3D. 
While it enables material selection from novel views with good consistency, each camera movement would require querying the model anew, with no guarantee of consistency in challenging cases on long frame sequences.
To lift our selection to 3D, we consolidate the 2D similarity maps of a sparse set of key-frames into a 3D similarity point cloud. Combined with nearest-neighbor queries, we can recover (and display) selections from any viewpoint on the 3D shape in a few milliseconds.
We show an overview of our method in  \Cref{fig:overview}.

\subsection{Fine-tuning for 2D material selection}

We re-purpose the SAM2 model \cite{ravi2024sam2} to material selection in the video domain. 
SAM2 uses an efficient \ac{ViT} image encoder \cite{ryali2023hiera} to produce a per-frame image embedding, and infers a per-pixel object similarity value for each frame. 
The key novel component in SAM2 is the memory attention module, which conditions the current frame embedding on the embeddings of past and future frames in the sequence, allowing the model to reason both spatially and temporally. 
These embeddings are then combined with the encoded conditioning query (\eg, a click on a pixel) in the mask decoder, producing per-frame similarity masks. 
As we will show in our experiments, correctly fine-tuning this memory module is key to achieving multiview-consistent selection results. 

\begin{wrapfigure}[18]{r}{0.26\textwidth}
    \vspace{-10pt}
    \centering
    \includegraphics[width=0.26\textwidth]{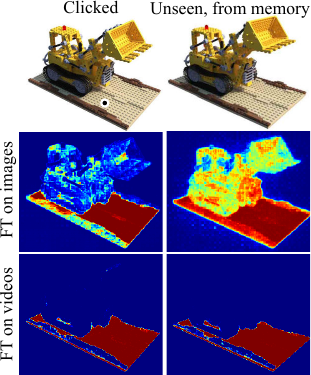}
    \vspace{-15pt}
    \caption{Fine-tuning on images creates artifacts on unseen frames, when selections are inferred from the model's memory (right column). 
    }
    \label{fig:video_finetuning}
\end{wrapfigure}

While our initial experiments confirmed SAM2's good cross-view consistency, they also revealed a tendency to select based on semantic function (\ie  by parts) instead of materials. 
We therefore fine-tune the model for the task of material selection.
Specifically, we freeze the encoder throughout our fine-tuning to preserve the rich priors learned from millions of images and tune the remainder of the architecture (see \cref{fig:sam_overview}). 
We find that training solely on \emph{image} data for material selection (\eg, the Materialistic dataset \cite{sharma2023materialistic}), performs reasonably well on clicked frames, but leads to a significant drop in cross-frame selection consistency, as shown in \cref{fig:video_finetuning}. 
We attribute this to the fact that on unseen frames, the model must infer the material selection from memory, but fine-tuning on images does not adapt the memory module since it is never queried for a single image. 

However, for 3D selection, cross-frame consistency is particularly important. 
We therefore design an object-centric \textit{video} dataset with material-segmentation annotation by randomly sampling objects, materials and environment maps, combining them into simple scenes containing one to a few objects.
We allow the same materials to appear multiple times and in different locations within a scene, to clearly disambiguate material and object selection.
We render 30 frames for each video using a random choice of four possible camera trajectories: zoom-in, zoom-out, spherical turntable and fly-over. 
Finally, to reduce the domain gap between natural and single-object images, we alpha-composite the environment map into the background (for additional dataset and training details, see Suppl. A.4.
\change{Since we freeze the pre-trained encoder to retrain its learned priors,} we find that 500 videos are sufficient to adapt the model to the material-selection domain. 

\setlength{\columnsep}{10pt}%

Our new video material dataset with dense per-frame material annotations lets us jointly fine-tune SAM2's memory attention module and the mask decoder. 
This way, we maintain multiview consistency while sig\-nifi\-cantly improving selections on unclicked frames (right column in \cref{fig:video_finetuning}). 
Our dataset and trained model are available \href{https://mfischer-ucl.github.io/sama/}{on the project page} and we show samples from the dataset in the supplemental.

\mycfigure{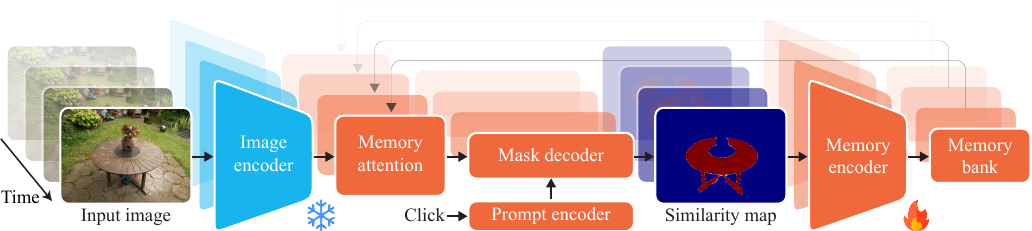}{Schematic overview of our fine-tuned model. The image encoder (in blue) is frozen, all other blocks (in red) are fine-tuned. Given an input image and a clicked pixel, the model outputs a material similarity map. Figure adapted from \citet{ravi2024sam2}.}


\subsection{Lifting 2D similarity to 3D}

Given a click on one image, our goal is to obtain a selection in 3D of all object parts that share the same material. 
A 3D selection is not only view-consistent by design, it also enables downstream applications (\eg, editing) that naturally operate in 3D on the object (surface). 
An entirely image-based pipeline would require running our 2D selection model for every new viewpoint, completely relying on the model's cross-frame consistency. Such a workflow would not be interactive (2--5\,sec per frame for simple selection visualization), would suffer from flickering due to residual multiview inconsistencies in long frame sequences, and would not be compatible with many downstream applications (\eg, mesh material replacement).
We therefore instead consolidate similarity maps from multiple viewpoints into a lightweight 3D similarity point cloud. From this cloud we can easily reconstruct (and display) a continuous 3D selection at interactive rates.

Given a click on the object, the initial camera, in which the click was performed, will serve to condition the memory module of our SAMa model, as it ensures that the material is not occluded.
We then render RGB and depth images from multiple viewpoints; for each RGB image we use our model to estimate the selection, based on the user-provided click, given the other images as video context through the memory bank. This process yields a per-viewpoint map representing the similarity to the user-clicked material. 
We project these 2D similarity maps to 3D using the previously rendered depth images, to obtain a 3D similarity point cloud located near the object surface.

Our approach works on any 3D representation that can be rendered from a given viewpoint and queried for depth. For NeRFs and 3D Gaussians we use a subset of the training views, while for meshes we use spherical Fibonacci sampling of camera positions pointed at the object's center. 
To ensure good performance of our fine-tuned video model, we arrange those views along a smooth trajectory via greedy iterative camera sorting described in Suppl. Algorithm 1.

For visualizing the 3D selection from novel views, we can reconstruct a continuous 3D similarity field via \ac{kNN} lookups into our previously constructed similarity point cloud. 
We use FAISS \cite{douze2024faiss} for performant large-scale, GPU-accelerated approximate nearest-neighbor queries~\cite{johnson2019billion} at the camera rays' 3D hitpoints. \change{As our point cloud contains one point per observed pixel, it is very dense (see Suppl. Figure 9 for an example).}

We cache and reuse the acceleration structure built by the library; we need to rebuild it only when the selected material changes.
With this approach, a new user selection from a novel viewpoint takes around 2\,sec (including 0.5\,sec for the structure construction), while querying the point cloud from a new viewpoint takes 10--20\,ms at 512--1024p image resolution. Suppl. Sec. A.2 provides additional details.

\begin{figure}[t]
    \centering
    \includegraphics*[width=\linewidth]{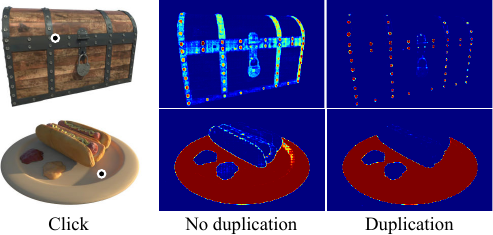}
    \vspace{-6mm}
    \caption{
        Effects of duplicating the clicked frame in the sequence. Similarity after frame duplication is significantly cleaner, as the model is forced to use the memory module.
    }%
    \vspace{-3mm}
    \label{fig:frame_duplication}
\end{figure}%

\subsection{Refinement}
\paragraph{Frame duplication.}

We observe that the frame where the user clicks exhibits significantly higher selection noise. This is due to the fact that, \change{by default, SAM2 does not} query the memory module for this selection, meaning that the model does not have access to the information in the other frames. To improve selection quality on this frame, we simply duplicate it. The first copy is used for conditioning the selection without memory module, and the second copy is included with the other frames in the sequence, using the memory module. We show the benefit of click-frame duplication in \cref{fig:frame_duplication} and on the original SAM2 model in the supplemental.

\paragraph{kNN-based voting.}

\begin{figure}[t]
    \centering
    \includegraphics*[width=\linewidth]{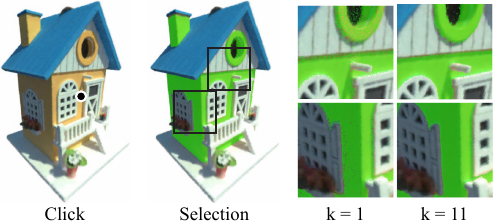}
    \vspace{-5mm}
    \caption{
        kNN 3D voting significantly reduces noise and improves selection quality, as seen from the insets.
    }%
    \vspace{-3mm}
    \label{fig:kNN_weighting}
\end{figure}

Thresholding our kNN-reconstructed 3D similarity field yields a binary selection field.
To ensure a clean selection, we use a binary voting scheme: we consider a 3D point as selected if more than half of its nearest neighbors pass the selection threshold. The threshold can be set by the user to adjust the selection, as in prior work~\cite{sharma2023materialistic}.
We show the effect of this aggregation strategy in \cref{fig:kNN_weighting}. 

\begin{figure}[t]
    \centering
    \includegraphics*[width=\linewidth]{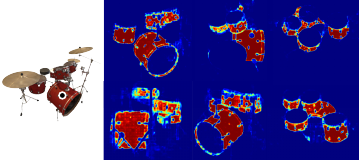}
    \vspace{-5mm}
    \caption{
        Conditioned by an initial click on the bass drum, our selection model achieves remarkable multiview-consistency in the presence of severe occlusions and perspective changes.
    }%
    \vspace{-3mm}
    \label{fig:selection_consistency}
\end{figure}%

\section{Evaluation}
\label{sec:results}


\subsection{Datasets}

We quantitatively evaluate our method on three datasets:
(1)~the eight scenes in the \ac{NeRF} dataset \cite{mildenhall2021nerf},
(2)~five real-world scenes from MIPNeRF-360 \cite{barron2022mip},
and (3) \change{our object-centric dataset test split containing 12 objects}.  
For synthetic assets we render material-ID maps which provide ground-truth annotations per view.
For real-world assets we hand-annotate five images per scene (see Suppl. Fig. 2). 

\change{We show additional qualitative evaluations on real-world captures and photogrammetry output in the supplemental.}

\subsection{Baselines}

We compare our method against three baselines. 
The first is the original Materialistic method \cite{sharma2023materialistic} for which we query for different views by re-projecting the initial click into the new view. 
In its default version, this can only be done for views where the original click is not occluded.
We can still query this baseline from new views thanks to our 3D-point-cloud lookup, but the results will be patchy as it cannot process all of the input views.
However, \citet{sharma2023materialistic} show that selection can work across two frames by computing the cross-attention between the initial click's $Q$ values and the $KV$ values of the other views. We extend this scheme to $n$ frames to process all unseen viewpoints, and refer to it below as ``Materialistic MV'' (multi-view).

Additionally, we compare against the multiview-consistent, but not material-aware, SAM2. 
Finally, ``Ours'' denotes our full method, including our fine-tuned network.
For all methods, we lift the results to 3D using our point-cloud representation. \change{We show  additional comparisons~\cite{kim_garfield_2024, guo2024semantic, li2024materialseg3d, cen2023segment} in Suppl. Figs. 11 -- 14.}



\subsection{Results}

We evaluate each method in 3D (after the point cloud lookup), along three axes: selection quality, robustness and multiview consistency. We report metrics on binary selection masks obtained by thresholding similarities against 0.5. Metrics are normed to $[0,1]$; see Suppl. Sec. B for details.

\newcommand{\best}[1]{$\mathbf{#1}$}
\begin{table}[]
    \centering
    \resizebox{\linewidth}{!}{%
        \begin{tabular}{@{}lrrrrrrrrrrrrr@{}}
\toprule
 Dataset &
  \multicolumn{2}{c}{NeRF \cite{mildenhall2021nerf}} &
  \multicolumn{2}{c}{MIPNeRF-360 \cite{barron2022mip}} &
  \multicolumn{2}{c}{Our Dataset} \\ \midrule

&
  \multicolumn{1}{c}{mIoU $\uparrow$} & \multicolumn{1}{c}{F1 $\uparrow$} & 
  \multicolumn{1}{c}{mIoU $\uparrow$} & \multicolumn{1}{c}{F1 $\uparrow$} & 
  \multicolumn{1}{c}{mIoU $\uparrow$} & \multicolumn{1}{c}{F1 $\uparrow$} \\  \cmidrule{2-7}

Ours &
  $\mathbf{0.48 \pm .2}$ & $\mathbf{0.58 \pm .3}$ & 
  $\mathbf{0.60 \pm .3}$  & $\mathbf{0.72 \pm .3}$  & 
  $\mathbf{0.69 \pm .2}$  & $\mathbf{0.78 \pm .2}$   \\
SAM2 &
  $0.33 \pm .2$ & $0.43 \pm .3$ & 
  $0.51 \pm .3$  & $0.65 \pm .3$  & 
  $0.36 \pm .2$  & $0.47 \pm .2$   \\
Mat. &
  $0.24 \pm .1$ & $0.36 \pm .2$ & 
  $0.31 \pm .3$  & $0.44 \pm .3$  & 
  $0.47 \pm .2$  & $0.59 \pm .2$    \\
Mat. MV &
  $0.27 \pm .2$ & $0.32 \pm .2$ & 
  $0.32 \pm .3$  & $0.47 \pm .3$  & 
  $0.51 \pm .2$  & $0.62 \pm .2$     \\
  \bottomrule
\end{tabular}%


    }
    \caption{
        Selection accuracy across datasets (columns) for several methods (rows), with 95\% confidence intervals. For the per-scene measurements and precision and recall, we refer to Suppl. Sec. B. Mat.\ is short for Materialistic \cite{sharma2023materialistic}. 
    }
    \label{table:selectionquality}
\end{table}

\begin{table}[]
    \centering
    \setlength{\tabcolsep}{16pt}
    \resizebox{\linewidth}{!}{%





\begin{tabular}{@{}l@{\hspace{0.95em}}lrrr@{}}
\toprule
\multirow{5}{*}{\rotatebox[origin=c]{90}{\makebox[30pt][r]{Consistency}}} & Dataset 

 & NeRF \cite{mildenhall2021nerf} 
 & MIPNeRF-360 \cite{barron2022mip} 
 & Our Dataset\\ \cmidrule(lr){2-5}

 & Ours &
  $\mathbf{2.2 \pm 0.2}$ & $1.4 \pm 0.2$ & $\mathbf{1.7 \pm 0.1}$  \\
 & SAM2 &
  $\mathbf{2.2 \pm 0.2}$ & $\mathbf{1.2 \pm 0.1}$ & $1.9 \pm 0.2$  \\
 & Materialistic &
  $5.5 \pm 0.3$ & $4.4 \pm 0.3$ & $5.9 \pm 0.4$  \\
 & Material. MV &
  $3.9 \pm 0.2$ & $4.1 \pm 0.4$ & $4.9 \pm 0.3$  \\ 
 \midrule 

\multirow{4}{*}{\rotatebox[origin=c]{90}{Robustness}} & Ours &
  $\mathbf{1.1 \pm 0.8}$ & $\mathbf{1.2 \pm 1.3}$ & $\mathbf{0.3 \pm 0.2}$  \\
 & SAM2 &
  $1.3 \pm 0.9$ & $2.9 \pm 3.8$ & $0.7 \pm 0.6$  \\
 & Materialistic &
  $3.2 \pm 0.6$ & $7.1 \pm 4.5$ & $1.8 \pm 1.0$  \\
 & Materialistic MV &
  $3.9 \pm 1.4$ & $3.5 \pm 1.5$ & $2.1 \pm 1.0$  \\ 
\bottomrule
\end{tabular}
    }
    \caption{
        Multiview consistency (top) and robustness (bottom) of our selection across unseen test views. We report Hamming distance ($\times 100$) with 95\% confidence intervals. Lower is better. 
    }
    \label{table:consistency_and_robustness}
\end{table}

\mycfigure{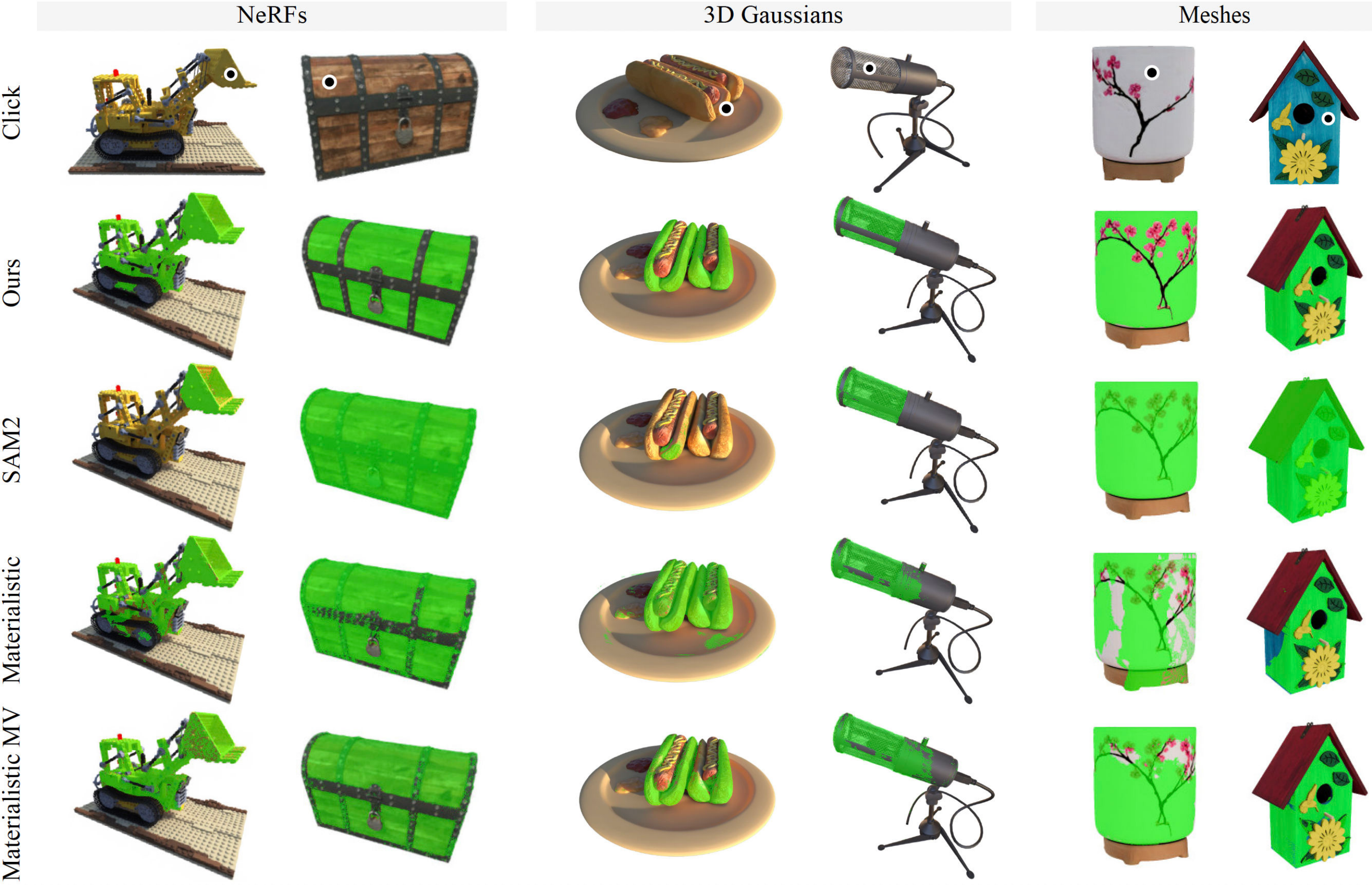}{Selection results on NeRFs, 3D Gaussians and meshes across objects (columns) and methods (rows). The top row shows the clicked view and the click, while selection results, from a novel view, are overlaid as green masks in the lower rows. We show results on assets from standard NeRF/3DGS benchmarks as well as a generated~\cite{wei2024meshlrm} mesh (cup) and photogrammetry capture (birdhouse), and more in supplemental material.}

\paragraph{Selection accuracy.}


We perform a click in one view, then for each of 50 random novel views we compare the obtained selection mask against a rendered ground-truth mask.
We compute \ac{mIoU}, a classical selection metric measuring the overlap between the masks. 
We also report F1 score which is the harmonic mean of precision and recall, and is more robust than either alone.
We average each metric over the views and over five random clicks on each material. 
We report the averages and 95\% confidence intervals across the datasets in \cref{table:selectionquality}, higher is better. 
Suppl. Sec. B provides a per-scene breakdown.

\paragraph{Multiview consistency.}


We demonstrate our method's multiview consistency in \cref{fig:selection_consistency}, and measure it numerically in \cref{table:consistency_and_robustness} top as follows. We perform a click in one view, then sample 50 novel views for which the clicked 3D point is unoccluded. For each view, we average the difference between the binary selection value at the point and the reference selection value of 1 in the clicked view.
Perfect multiview consistency means zero average difference, \ie all values are 1. 
Note that this metric does not quantify mask correctness; if a returned mask is wrong, but consistently so, the reported score will still be high. Both our method and SAM2 show similar consistency, while both Materialistic baselines, which do not benefit from the cross-frame memory mechanism, achieve lower consistency scores. 
\vspace{-2mm}
\paragraph{Robustness to click location.}


On a random view, we perform 5 random clicks within a single material. We then average the pairwise Hamming distances between the 5 selection masks. We report results in \cref{table:consistency_and_robustness} bottom, lower is better. 
Like the multiview metric above, this metric does not quantify mask correctness. We see that architectures that benefit from multi-frame context show better robustness.
\vspace{-2mm}
\paragraph{Qualitative evaluation.}

\begin{figure}[t]
    \centering
    \includegraphics*[width=\linewidth]{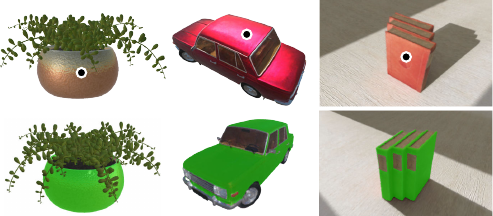}
    \vspace{-5mm}
    \caption{
        Our method is robust w.r.t.\ shading variations on the surface, shown here for reflections, specularity and shadows.
    }%
    \vspace{-3mm}
    \label{fig:shading_robustness}
\end{figure}

We show selection results on all our evaluated modalities (NeRFs, 3D Gaussians, meshes) in \cref{fig:results_qualitative}. 
We see that Materialistic and Materialistic MV do not work well on high-frequency boundaries of objects and that SAM2 cannot be trivially used for material selection, with only parts or entire objects selected at once.
In contrast, our method creates sharp boundaries, including around thin elements, and is robust to lighting variations (see \cref{fig:shading_robustness}).
For additional results on generated- as well as real-world meshes and neural assets, please see the supplemental material. 
We also include comparisons to MatSeg3D~\cite{li2024materialseg3d}, which focuses on mesh segmentation into semantic material classes, and methods focusing on object segmentation and selection in 3DGS \cite{guo2024semantic,zhou2024feature, kim_garfield_2024}.

We further evaluate our results in 2D without the point cloud lookup, which improves average mIoU by around 5\%. However, our 3D aggregation in a point cloud lookup provides a significant advantage in efficiency, reducing per-frame inference processing times from around 5\,sec in 2D to around 10\,ms in 3D (500$\times$ faster), making it a more practical choice overall. This difference in quality is mainly explained by the depth estimation quality on volumetric representations, which is not always perfect.

\section{Applications}


\subsection{Segmentation}
\label{sec:segmentation}

While our method targets selection, we propose an automatic segmentation mode to divide an object into material subparts inspired by the image-level sampling in SAM~\cite{sam}.


\change{Densely sampling an entire object from multiple views is impractical (500-click sampling of the Lego asset in \cref{fig:video_finetuning} takes $\sim$20\,min). 
Instead, we propose a light-weight, greedy alternative. We first randomly sample a click on the asset, for which we compute the material similarity (as outlined in \cref{sec:method}) in a fixed set of 15 views (elevation in $\left[-30\degree, 0\degree, 30\degree\right]$, azimuth in five 72\textdegree\, increments). We binarize these per-view selections and project them to UV space, assigning a unique ID to the clicked material. We then sample a new point in an area that has not yet been segmented.}

\begin{figure}[t]
    \centering
    \includegraphics*[width=\linewidth]{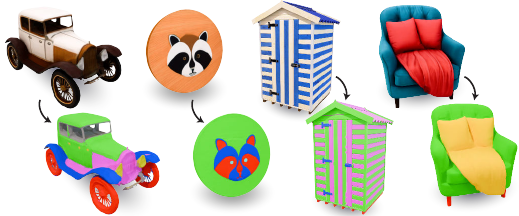}
    \vspace{-5mm}
    \caption{
        Automatic segmentation of synthetic ($3^{rd}$) and generated \cite{wei2024meshlrm} ($1^{st}$, $2^{nd}$, $4^{th}$) meshes into material ID maps.
    }%
    \vspace{-3mm}
    \label{fig:mesh_segmentation}
\end{figure}

\change{We repeat this process until the entire mesh surface is segmented, which yields an ID map in UV space in which each point of the visible mesh surface corresponds to a unique material ID via its UV coordinate.}
\change{We can use this map to edit specific material regions, for PBR replacement, or to segment the mesh along the material ID regions. 
With this approach, the segmentation of a complete mesh takes under 10 seconds. 
We show results of our method in \cref{fig:mesh_segmentation}.}



\subsection{Editing}

Using our material selection results, we can easily edit the selected regions. We show various edits and applications for NeRFs, 3D Gaussians and meshes in \cref{fig:Teaser}.

\textbf{NeRFs.}
For NeRFs, we demonstrate color editing. We ray-march the NeRF as usual, but for each 3D point we query whether it has been selected.
If yes, we adjust the color returned by the NeRF through a color shift in LAB space, to preserve relative shading and lighting information.
We show an example in \cref{fig:Teaser} and in Suppl. Fig. 7.

\textbf{Material-aware 3D Gaussians.}
For 3D Gaussians, we use our material segmentation step (described in \cref{sec:segmentation}). We then render the respective material masks for each training view and convert them to ID masks, so each pixel in the training images is associated with a material index. We then re-train the Gaussians with an extra channel for materials which is treated like the RGB channels for rasterization. This creates a clear separation between Gaussians at material boundaries, simplifying downstream edits, and provides a per-Gaussian material handle. We can now select all Gaussians that encode, \eg, material number two, and edit their properties such as color, position or density. In \cref{fig:Teaser} we move the gazebo's wooden base upwards and set the white painted regions' density to zero.

\textbf{Meshes.}
For selection on meshes, we exploit their UV parametrization by writing the selected material similarities to a 2D UV map.
This enables trivial creation of material-ID maps, or change of a selected material. 
Here, because the similarities are directly projected to pixel values, we find it beneficial to use the hole-filling and sprinkle-removal techniques described in the original SAM2 paper \cite{ravi2024sam2}. 

We show results on the output of text-to-mesh generated assets \cite{wei2024meshlrm}. Using our automatic segmentation we can easily replace the diffuse textures on a text-to-3D generated asset with PBR materials (see Suppl. Fig. 10).

\section*{Future work}

We find our method to significantly improve material selection in 3D. However, some limitations remain to be addressed; for instance, selecting materials on objects like glass and mirrors remains challenging as it is unclear if a user would prefer to select the transparent/mirror material or what is behind/reflected. 
Similarly, our method currently assumes a fixed definition of materials and cannot adapt to user-defined preferences, such as focusing solely on specific parameters like roughness. 

Further, SAMa depends on precise 3D reconstruction for accurate material selection. Errors in depth reconstruction can cause noise in our point clouds and inaccurate lookups in novel views, an issue which can be mitigated by improving depth estimation in volumetric reconstruction \cite{zhang2024rade}. 
Finally, unseen parts of meshes are selected based on their nearest visible 3D points, which could be improved with a UV space propagation of the selection~\cite{knn_matting}.

\section*{Conclusion}
\label{sec:conclusion}

We present SAMa, a material selection model for 3D, leveraging a video model for cross-view consistency and a simple yet efficient projection to 3D in the form of a similarity point cloud. Our approach enables interactive material selection, visualization and downstream manipulation of the 3D assets at high quality. As we specialize the SAM2 video model to a new modality, we find that video-based finetuning is crucial, and that 500 varied videos are sufficient to change the modality. We believe this opens interesting opportunities to explore selection across various modalities.


{
    \small
    \bibliographystyle{ieeenat_fullname}
    \bibliography{main}
}


\end{document}


\definecolor{darkgreen}{RGB}{0,110,0}
\definecolor{darkred}{RGB}{170,0,0}
\def\greencheckmark{\textcolor{darkgreen}{\checkmark}}
\def\redxmark{\textcolor{darkred}{\text{\ding{55}}}}  %

\addeditor{valentin}{VD}{0.7, 0.0, 0.7}
\addeditor{vde}{VD}{0.7, 0.0, 0.7}
\addeditor{thibault}{TG}{0.0, 0.0, 0.8}
\addeditor{vova}{VK}{0.0, 0.5, 0.0}
\addeditor{michael}{MF}{0.1, 0.5, 0.9}
\addeditor{iliyan}{IG}{0.8, 0.0, 0.0}
\addeditor{tobias}{TR}{0.2, 0.8, 0.1}

\showeditsfalse
\showeditstrue



\maketitlesupplementary
\appendix


In this supplemental document we provide additional details on training and implementation, as well as results that could not be included in the main text due to space restrictions. We strongly encourage the reader to view the videos in our supplemental HTML material for 3D selection visualizations, examples of our fine-tuning material dataset, and a video of our application GUI.

\section{Implementation details}

\subsection{Fine-tuning}
\label{sec:suppl_finetuning}

As mentioned in paper \cref{sec:method}, we fine-tune parts of the SAM2 \cite{ravi2024sam2} model on material-specific video data. 
For all our experiments, we use the model in its ``large'' configuration, employing the Hiera \cite{ryali2023hiera} image encoder with ca.\ 212M params, which yielded the best results in our experiments.

As the original SA-V \cite{ravi2024sam2} dataset, we encode our video dataset as MP4  videos with 1024$\times$1024 resolution and the annotations in CoCoRLE encoding for efficient storage. 

Our video dataset sub-samples the video by skipping every other video frame to increase the intra-frame distance, and then randomly chooses sequences of six consecutive sub-sampled frames. 
For each material and each frame, we sample a click. We do not select a material if it is barely visible in the frames, \ie, if it occupies less than 0.02\% of the frame (150 pixels). 
We erode the material's ground-truth mask before using it as a sampling mask, ensuring that the sampled click is at least four pixels away from the material's border. 
We sample a positive click with 80\% probability, and a negative click on a random other material with 20\% and reverse the temporal order of the frame sequence with a chance of 50\%. 
During the forward pass of the model, we use every other frame as a clicked frame and thus force the model to use its memory attention module to infer the selection for the intermediate, unclicked frames. 
Additionally, we make a random 50\% choice between sampling the most salient material in the frame (with the highest number of annotated pixels) and any other material. 

\mycfigure{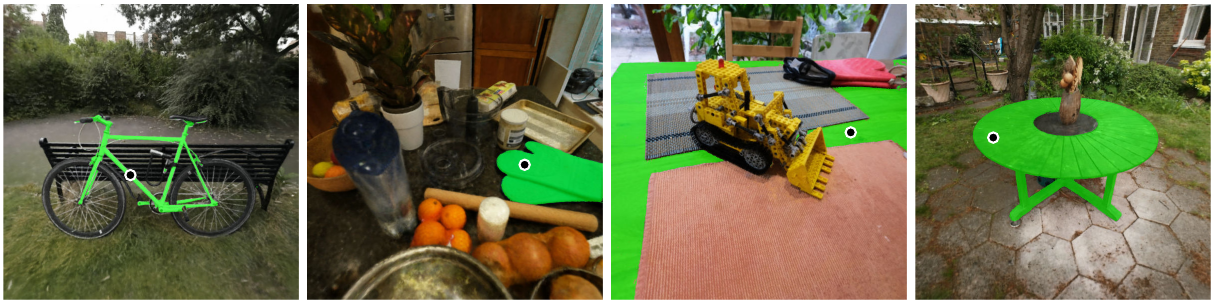}{Selection results on real-world scenes from the MIPNeRF360 dataset \cite{barron2022mip}.}

During training, we compute the per-frame loss on the model prediction and ground-truth annotation via the sum of two losses, a binary cross-entropy followed by a sigmoid (using the log-sum-exp \cite{bishop2006pattern} trick for numerical stability) and a sigmoid-normalized Dice loss \cite{milletari2016v} to account for the imbalance between (large) background and (smaller) material masks. 
We use the AdamW optimizer with weight decay $0.01$ and learning rate $1\times10^{-5}$. 

We additionally experiment with mixed video- and image-finetuning and find that the results perform roughly on-par with our video model when training on our video-dataset and 20\% of the Materialistic \cite{sharma2023materialistic} data set mixed in. 
For simplicity, all results in the main text therefore use solely our video-finetuned model.

\subsection{kNN lookup}
\label{sec:suppl_knn}

As explained in the main text, we perform \ac{kNN} lookup into our similarity point cloud to infer the material selection for new, unseen views. 
Here, we take advantage of modern, GPU-accelerated large-scale queries via the FAISS library \cite{douze2024faiss, johnson2019billion}. 

Specifically, we use the \textsc{IndexFlatL2} index for exact search w.r.t. the points' $\textrm{L}_2$ distance, encoded as an \textsc{IndexIVFFlat} for compactness, with 100 clusters, and push it to the GPU (a cluster is a representative subset of the data that can be traversed efficiently and narrows down the search region during later query operations). 
This index, as mentioned in the main text, must be re-constructed after each new click, since the initial camera from which the click was performed will add to, and therefore change, the similarity point cloud. 
This re-construction takes around 0.5 seconds (all timings, including those in the main text, are reported on a single NVIDIA 40GB A100).

Once the index is built, we visit five clusters during the search for the top-k nearest neighbors.
We found this number of visited clusters to be a hyperparameter which, even with the lowest setting of a single cluster, does not significantly deteriorate performance since the point cloud is relatively dense. \change{We show an example of a typical point cloud in Figure 9. Thanks to the point cloud density our selection handles sharp edges well}.

\subsection{Camera subsampling}
\label{sec:camera_sampling}

To infer the 2D similarities which will later be projected to 3D, we need to sub-sample a set of cameras that cover the object well. 
Recall that we fine-tuned SAM2 with our material-centric video dataset using four different camera trajectories, all with smooth view progression, while now we additionally need to ensure maximum object coverage. 
For NeRFs and 3D Gaussians, we thus sub-sample 20\% of the training views, for meshes we use spherical Fibonacci sampling with 30 sampled cameras. 
Once we have sub-sampled the cameras, we need to sort them into a coherent, smooth trajectory to enable our video model to keep temporal consistency between the frames. 
We use a greedy iterative search to achieve a smooth trajectory from the initial camera, as detailed in \cref{alg:camerasampling}.

\subsection{Dataset details}
\label{sec:suppl_dataset}
To construct our fine-tuning dataset, we procedurally generate short multi-object videos using randomly picked objects from a subset of 9,082 textured Substance 3D objects. 
For each video, we randomly sample at least two objects and place them into a shared scene. 
Objects are randomly displaced by up to half of their bounding box extent to reduce large spatial overlap. 
Objects consisting of a single material are excluded, as they do not provide meaningful supervision for material selection.

We assign materials by sampling from a library of 29,472 Substance material maps, including multiple realizations of the same base material (\eg, different variants of the same type of wood). 
Materials are assigned to object parts such that they appear at least twice within a scene to ensure sufficient positive supervision and disambiguate from object selection. 
Material assignments remain consistent throughout a video but are resampled independently across videos. 
We generate dense per-pixel material annotations using integer indices, reserving the label 0 for the background.

Rendering is performed in Blender 4.3, using the Principled BSDF shader for all materials. 
To illuminate the scene, we randomly select an HDR environment map from a set of 420 HDRIs sourced from PolyHaven. 
In order to reduce the domain gap between our synthetic and real-world videos, we alpha-composite the rendered objects onto their envmap background.

We generate videos using four camera motion patterns: turntable, flyover, zoom-in, and zoom-out. 
Spherical, fixed-radius turntable trajectories are sampled with 33\% probability since they represent the dominant camera trajectories (after sorting) that are used during SAMa's inference phase. 
Flyover trajectories are also sampled with 33\% probability, as they resemble realistic camera motion and are most similar to those found in the SA-V \cite{ravi2024sam2} dataset. 
Zoom-in trajectories are sampled with 22\% probability since during the zoom-in phase, more detail becomes visible, and zoom-out trajectories account for the remaining 12\%. 
We leave experimenting with this percentages and more camera trajectories to future work.

\begin{algorithm}[b]
    \caption{
        Camera trajectory sorting, starting from an initial camera. \textsc{CalcNorms} calculates the spatio-angular distances between a given camera and all other cameras.
    }
    \begin{algorithmic}[1]
        \Statex \textbf{Input:} initial camera $i$, other cameras $o$
        \Statex \textbf{Output:} sorted cameras
        \Procedure{SampleCameraTrajectory}{}
            \State $\textrm{curr} \gets i$ \Comment{set current camera}
            \State $\textrm{sorted} \gets \left[ \, \textrm{curr} \, \right]$ \Comment{initialize sorted cameras list}
            \While{$\textrm{len}(o) > 0$}
                \State $\textrm{norms} \gets \Call{CalcNorms}{\textrm{curr}, o}$
                \State $\textrm{cidx} \gets \textrm{argmin\,(norms)}$ \Comment{closest to current}
                \State sorted.append($o\left[\textrm{cidx}\right]$)
                \State curr $\gets$ $o\left[\textrm{cidx}\right]$
                \State $o\left[\textrm{cidx}\right]$.pop()
            \EndWhile
            \State \Return sorted
        \EndProcedure
    \end{algorithmic}
    \label{alg:camerasampling}
\end{algorithm}

\section{Additional quantitative results}
\label{sec:suppl_metrics}

\mycfigure{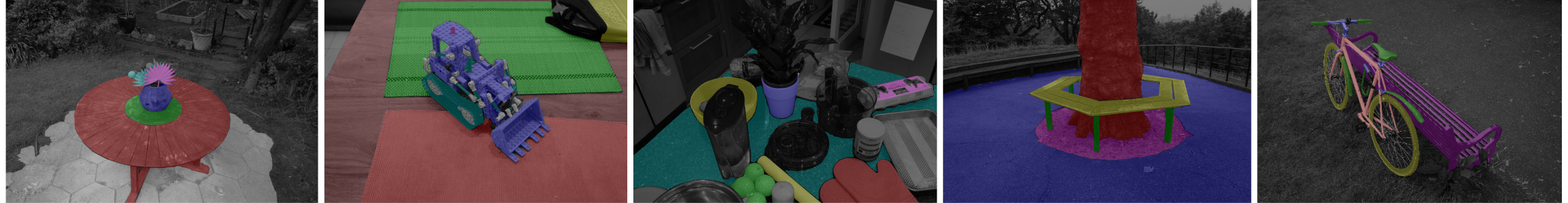}{Exemplary visualizations of our annotated test frames from the MIPNeRF360 dataset \cite{barron2022mip}.}

\mycfigure{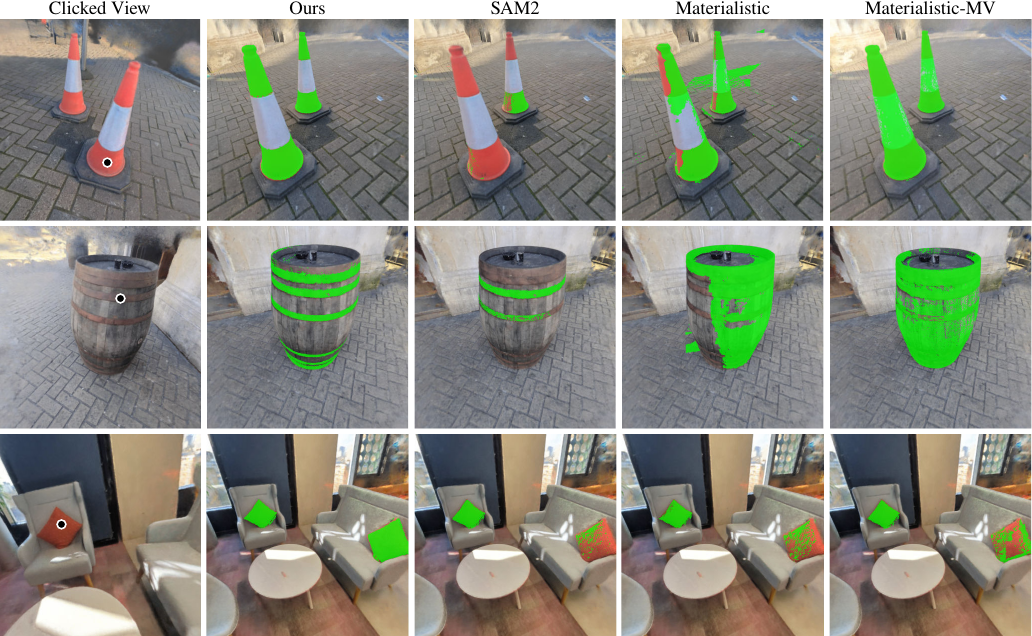}{We test our method and competitors on 3DGS scenes we capture and find that it generalizes and performs considerably better than existing approaches. 
The first column shows the clicked view, the subsequent columns show the selection result from a novel, unseen view.}

We here report a more detailed, per-scene evaluation of the metrics reported in the main text.
The per-scene measurements for robustness and multiview-consistency are in \cref{tab:robustness_suppl} and \cref{tab:mvconsistency_suppl}, respectively. 

Additionally, we report the per-scene selection accuracy as \ac{mIoU} and F1 scores.
F1 is more robust than precision or recall alone, since either individual metric can easily be gamed by failure cases. 
Precision quantifies the relevance of the selected data (when the model says material A, is it really material A?), and can therefore easily be cheated by simply selecting a small amount of high-confidence elements (\eg, in our case, just the clicked pixel). 
Recall quantifies the amount of returned relevant data (when there is material A, how much of it does the model find?), and can easily be deceived by always selecting all the elements (\eg, in our case, a mask full of 1's). 
We show both \ac{mIoU} and F1, computed on the NeRF-, MIPNeRF360- and our dataset, in \cref{tab:qualitative_suppl_nerf}, \cref{tab:qualitative_suppl_mipnerf} and \cref{tab:qualitative_suppl_ourdataset}, respectively.
We perform the evaluation on 3D Gaussians for rendering speed.
For the real-world scenes from the MIPNeRF dataset, we found the Gaussian's depth to not be sufficiently accurate and therefore use NeRFacto \cite{tancik2023nerfstudio}.

The quantitative evaluation confirms our qualitative findings: our method consistently performs well for the task of material selection, beating the other baselines in the majority of cases. 
In select cases, for instance the \textsc{Mic} scene from the NeRF dataset (see \cref{tab:qualitative_suppl_nerf}), SAM2 wins in terms of selection accuracy, since the materials of the object are visually indistinguishable from one another and applied to the object's subparts, which have a tendency to be selected by SAM2. 
Both Materialistic-based baselines under-perform in all experiments. 
This can be attributed in part to the fact that they are not multiview consistent, but, equally important, to the fact that the underlying model generally attends to coarser structures (due to the different ViT patchsizes, see \cref{fig:similarity_2D}) and is not sufficiently sensitive to object (sub-)parts. 

\myfigure{similarity_2D}{2D selection results of the different methods for various models. We do not perform any point cloud lookup or novel view inference, the shown heatmap is obtained by directly feeding the clicked frame to the model.}

\section{Additional qualitative results}
\label{sec:suppl_visuals}

\myfigure{robustness_suppl}{Robustness of the different approaches (rows) for clicks on different locations of the same material (columns).}

\myfigure{frame_duplication_suppl}{The effects of our frame duplication strategy translate from our SAMa model to the original SAM2 model.}

We show additional examples of recoloring NeRFs based on our material-selection in \cref{fig:nerf_colorediting_suppl}. 

We show examples of our hand-annotated frames from the MIPNeRF dataset which we used for evaluation in \cref{fig:mipnerf_annotated}. 
Additionally, we show examples of material selection on real-world scenes from these MIPNeRF360 scenes \cite{barron2022mip} in \cref{fig:mipnerf_results}.

\myfigure{nerf_colorediting_suppl}{Additional examples of editing the NeRF's color based on the user's selected material.}

\begin{figure}
    \centering
    \includegraphics[width=\linewidth]{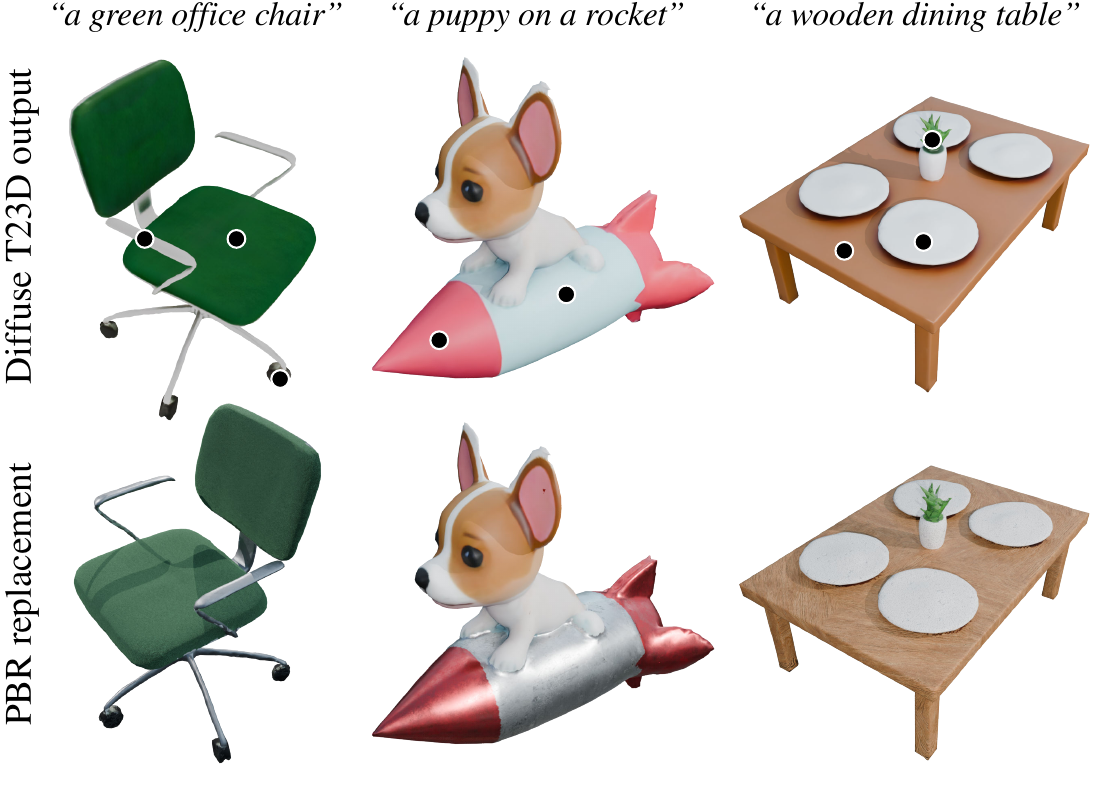}
    \vspace{-5mm}
    \caption{Our method can be used to select, and subsequently replace, the diffuse-only materials commonly found on text-to-3D~\cite{wei2024meshlrm} pipeline output meshes with PBR materials.}
    \vspace{-5mm}
    \label{fig:pbr_replacement}
\end{figure}

\myfigure{selection_and_pc_viz}{Visualization of the selection point cloud (pre-computed by SAMa, right subfigure, green \& brown) and the query points obtained from a novel view (used for the look-up into the selection point cloud). The red inset's query points are visualized in blue in the right sub-figure. Best viewed zoomed-in.} 

As claimed in the main text, our frame duplication strategy not only improves SAMa's predictions, but also helps to improve prediction confidence on the original SAM2 architecture, which we visualize in \cref{fig:frame_duplication_suppl}.

To add to our robustness evaluation, we show a qualitative example of how robust the methods are to different clicks on the same material in \cref{fig:robustness_suppl}.

We also show the 2D material selection accuracy for all models in \cref{fig:similarity_2D}. 
From this figure, it becomes evident that the SAM-based methods benefit significantly from the smaller patchsize of the image encoder: Hiera, the encoder used by the SAM2 architecture (Ours, SAM2) uses a four-times smaller patchsize of $4\times4$, whereas Materialistic-based methods employ DINO features, which use a patchsize of $8\times8$, resulting in blurrier edges.
We would like to emphasize that the input resolution is the same for all models, 512p. 
Moreover, we observe that our model deals well with perspective distortion (middle row in \cref{fig:similarity_2D}) and low-contrast input (bottom row in \cref{fig:similarity_2D}). 
Finally, we show thumbnail renderings of our synthetic dataset in \cref{fig:dataset_thumbnails}.

\myfigure{photogrammetry_results}{We compare our method and competitors on real mesh data that was obtained by using photogrammetry on real-world objects with Polycam. For animated version of these results, see the electronic supplemental.} 

\section{Additional comparisons}
\label{sec:suppl_comparisons}

We qualitatively compare our method against other 3D-aware selection and segmentation methods.
Note that a full quantitative comparison against these baselines is not feasible due to their long per-asset optimization times. 

\noindent \textbf{Selection.} In \cref{fig:garfield}, we show a comparison against Garfield \cite{kim_garfield_2024}, which requires asset-specific pre-training and does not target materials. 
The same holds for Feature-3DGS \cite{zhou2024feature} (right side in \cref{fig:lifting_methods}).
In contrast, our approach works with arbitrary assets and \emph{without} asset-specific pre-training, as it merely needs to render the existing 3D asset to images and back-project the obtained similarity values. 
Our times from click-to-selection are therefore around three orders of magnitude lower. 
\change{Finally, the comparison to SA3D~\cite{cen2023segment} in \Cref{fig:sa3d_comparison} shows that our approach correctly selects materials for NeRFs as well, at much lower runtimes than SA3D.}

\myfigure{garfield}{Comparison to Garfield \cite{kim_garfield_2024}, which cannot be run without asset-specific pre-training and does not target material selection.}

\myfigure{sa3d_comparison}{Comparison against SA3D. SAMa quickly selects the material, while SA3D, after lengthy pre-optimization, selects object (sub-) parts.}

\noindent \textbf{Segmentation.} Additionally, in \cref{fig:lifting_methods}, we show a comparison against SemanticGaussians \cite{guo2024semantic} (left side), who inject semantic knowledge into a 3DGS capture via a separately trained network, but whose predictions are coarse, not material-aware and, by design, limited to 3DGS assets. 
Similarly, \cref{fig:segmentation_sameclrs} shows an evaluation of MaterialSeg3D \cite{li2024materialseg3d}, a 3D segmentation method for materials that works on meshes. 
While this methods works well and is multiview-consistent, it classifies materials into 14 predefined semantic classes (metal, wood, plastic, ...) and thus is not able to distinguish between different materials within a category, such as the different types of wood on the beach hut.
Our method, in contrast, selects and segments these materials correctly. 

\change{In summary, SAMa performs well on material-aware selection and segmentation in 3D, works on NeRFs, 3DGS and meshes, provides interactive click-to-selection and can infer selection results for novel views in real-time.}

\myfigure{segmentation_sameclrs}{Comparison between the material-aware segmentation of SAMa and the semantic segmentation of MaterialSeg3D \cite{li2024materialseg3d}
} 

\myfigure{lifting_methods}{Performance and timing of 2D-to-3D lifting methods.
SemanticGaussians \cite{guo2024semantic} does not target materials, while Feature-3DGS \cite{zhou2024feature} requires a lengthy pre-optimization.}


\begin{table*}[!h]
    \centering
    \resizebox{\textwidth}{!}{%
        \begin{tabular}{lrrrrrrrrrrrrrrrrrrrrrrrrr}\toprule
&\multicolumn{2}{c}{\textsc{WChair}} &\multicolumn{2}{c}{\textsc{Coffee}} &\multicolumn{2}{c}{\textsc{Perfume}} &\multicolumn{2}{c}{\textsc{Chest}} &\multicolumn{2}{c}{\textsc{Couch}} &\multicolumn{2}{c}{\textsc{Bike}} &\multicolumn{2}{c}{\textsc{Hut}} &\multicolumn{2}{c}{\textsc{Burger}} &\multicolumn{2}{c}{\textsc{Plant}} &\multicolumn{2}{c}{\textsc{Postbox}} &\multicolumn{2}{c}{\textsc{Car}} &\multicolumn{2}{c}{\textsc{Pooltable}} \\\cmidrule{2-25}
&mIoU &F1 &mIoU &F1 &mIoU &F1 &mIoU &F1 &mIoU &F1 &mIoU &F1 &mIoU &F1 &mIoU &F1 &mIoU &F1 &mIoU &F1 &mIoU &F1 &mIoU &F1 \\\cmidrule{2-25}
Ours &0.73 &0.84 &0.91 &0.95 &0.92 &0.96 &0.82 &0.90 &0.41 &0.56 &0.71 &0.83 &0.79 &0.88 &0.89 &0.94 &0.79 &0.88 &0.94 &0.97 &0.34 &0.51 &0.57 & 0.73 \\
SAM2 &0.44 &0.60 &0.47 &0.64 &0.91 &0.96 &0.41 &0.56 &0.09 &0.16 &0.22 &0.36 &0.17 &0.25 &0.57 &0.72 &0.26 &0.41 &0.69 &0.81 &0.06 &0.11 &0.12 & 0.21 \\
Materialistic &0.51 &0.67 &0.51 &0.68 &0.81 &0.89 &0.46 &0.61 &0.18 &0.30 &0.63 &0.77 &0.44 &0.61 &0.25 &0.40 &0.74 &0.85 &0.91 &0.95 &0.11 &0.19 &0.15 & 0.26 \\
Materialistic MV &0.61 &0.75 &0.48 &0.64 &0.89 &0.94 &0.51 &0.65 &0.17 &0.29 &0.57 &0.73 &0.52 &0.69 &0.53 &0.67 &0.75 &0.86 &0.94 &0.97 &0.10 &0.18 &0.25 & 0.40 \\
\bottomrule
\end{tabular}

    }
    \caption{
        Per-scene metrics on our synthetic dataset for the different scenes (columns) and methods (rows). Higher is better.
    }
    \label{tab:qualitative_suppl_ourdataset}
\end{table*}

\begin{table*}[!htp]
\centering
\resizebox{\textwidth}{!}{%
\begin{tabular}{lrrrrrrrrrrrrrr}\toprule
&\textsc{Lego} &\textsc{Hotdog} &\textsc{Ship} &\textsc{Ficus} &\textsc{Mic} &\textsc{Drums} &\textsc{Materials} &\textsc{Chair} &\textsc{Garden} &\textsc{Kitchen} &\textsc{Counter} &\textsc{Treehill} &\textsc{Bicycle} \\\cmidrule{2-14}
Ours &0.21 &1.01 &1.68 &0.45 &2.89 &0.85 &1.47 &0.25 &0.38 &0.19 &1.25 &2.79 &1.22 \\
SAM2 &0.43 &1.04 &2.55 &0.46 &1.75 &0.43 &0.33 &3.03 &0.76 &4.41 &0.22 &1.51 &7.68 \\
Materialistic &4.33 &2.69 &2.62 &2.40 &3.10 &2.48 &3.69 &3.88 &10.24 &6.16 &13.51 &4.83 &5.90 \\
Materialistic MV &7.37 &3.17 &2.33 &2.99 &4.38 &2.51 &3.67 &4.82 &3.27 &2.35 &5.04 &4.47 &2.52 \\\cmidrule{1-14}
&&\textsc{WChair} &\textsc{Coffee} &\textsc{Perfume} &\textsc{Chest} &\textsc{Couch} &\textsc{Bike} &\textsc{Hut} &\textsc{Burger} &\textsc{Plant} &\textsc{Postbox} &\textsc{Car} &\textsc{Pooltable} \\ \cmidrule{3-14}
Ours &&0.06 &0.09 &0.01 &0.12 &0.60 &0.95 &0.87 &0.04 &0.61 &0.15 &0.43 &0.15 \\
SAM2 &&0.10 &0.01 &0.51 &0.02 &0.34 &0.45 &2.25 &0.60 &0.02 &3.31 &1.17 &0.05 \\
Materialistic &&0.42 &0.73 &0.50 &1.28 &4.80 &2.53 &2.63 &1.16 &0.60 &0.71 &1.86 &4.88 \\
Materialistic MV &&0.19 &0.85 &0.61 &2.30 &4.83 &2.34 &5.46 &2.97 &2.35 &0.68 &1.25 &1.95 \\
\bottomrule
\end{tabular}
}
\caption{Per-scene (columns) breakdown of our robustness evaluation metric for all methods (rows) from the main text. Lower is better. The NeRF- and MIPNeRF360-scenes are in the top sub-table, our custom scenes in the bottom sub-table. This only evaluates the robustness and not whether the selection is correct.}\label{tab:robustness_suppl}
\end{table*}

\begin{table*}[!htp]
    \centering
    \resizebox{\textwidth}{!}{%
        \begin{tabular}{lrrrrrrrrrrrrrr}\toprule
&\textsc{Lego} &\textsc{Hotdog} &\textsc{Ship} &\textsc{Ficus} &\textsc{Mic} &\textsc{Drums} &\textsc{Materials} &\textsc{Chair} &\textsc{Garden} &\textsc{Kitchen} &\textsc{Counter} &\textsc{Treehill} &\textsc{Bicycle} \\\cmidrule{2-14}
Ours &0.91 &1.20 &2.20 &0.30 &5.64 &0.62 &5.77 &0.85 &0.18 &0.72 &0.87 &1.32 &3.71 \\
SAM2 &2.99 &1.44 &3.03 &0.37 &1.46 &0.94 &1.54 &6.13 &0.28 &1.04 &0.40 &1.43 &2.77 \\
Materialistic &5.18 &4.57 &7.98 &0.62 &4.59 &1.77 &12.59 &6.56 &2.59 &4.40 &2.29 &9.06 &5.92 \\
Materialistic MV &2.79 &4.87 &2.97 &0.45 &4.26 &1.00 &8.58 &6.32 &2.15 &2.11 &0.89 &9.00 &6.36 \\ \midrule
&&\textsc{WChair} &\textsc{Coffee} &\textsc{Perfume} &\textsc{Chest} &\textsc{Couch} &\textsc{Bike} &\textsc{Hut} &\textsc{Burger} &\textsc{Plant} &\textsc{Postbox} &\textsc{Car} &\textsc{Pooltable} \\ \cmidrule{3-14}
Ours & &0.89 &0.31 &0.26 &1.16 &1.05 &1.17 &3.46 &0.32 &0.61 &0.69 &1.61 &9.03 \\
SAM2 & &1.60 &1.61 &0.47 &4.03 &1.98 &5.58 &16.63 &1.54 &0.68 &2.69 &2.88 &20.95 \\
Materialistic & &3.12 &2.72 &0.73 &8.12 &3.61 &2.92 &13.27 &7.32 &2.33 &0.93 &13.79 &10.66 \\
Materialistic MV & &1.78 &3.11 &0.30 &6.79 &2.08 &2.32 &11.04 &4.07 &1.90 &0.61 &16.89 &5.40 \\
\bottomrule
\end{tabular}
    }
    \caption{
        Per-scene (columns) breakdown of our multiview-consistency evaluation metric for all methods (rows) from the main text. Lower is better. The NeRF- and MIPNeRF360-scenes are in the top sub-table, our custom scenes in the bottom sub-table. This only evaluates the multiview-consistency and not whether the selection is correct.
    }
    \label{tab:mvconsistency_suppl}
\end{table*}

\begin{table*}[!htp]
    \centering
    \resizebox{\textwidth}{!}{%
        \begin{tabular}{lrrrrrrrrrrrrrrrrr}\toprule
&\multicolumn{2}{c}{\textsc{Lego}} &\multicolumn{2}{c}{\textsc{Hotdog}} &\multicolumn{2}{c}{\textsc{Ship}} &\multicolumn{2}{c}{\textsc{Ficus}} &\multicolumn{2}{c}{\textsc{Mic}} &\multicolumn{2}{c}{\textsc{Drums}} &\multicolumn{2}{c}{\textsc{Materials}} &\multicolumn{2}{c}{\textsc{Chair}} \\ \cmidrule{2-17}
&mIoU &F1 &mIoU &F1 &mIoU &F1 &mIoU &F1 &mIoU &F1 &mIoU &F1 &mIoU &F1 &mIoU &F1 \\ \cmidrule{2-17}
Ours &0.78 &0.87 &0.87 &0.93 &0.06 &0.12 &0.68 &0.81 &0.24 &0.39 &0.25 &0.39 &0.16 &0.27 &0.76 &0.87 \\
SAM2 &0.05 &0.09 &0.77 &0.87 &0.10 &0.18 &0.68 &0.81 &0.51 &0.68 &0.07 &0.14 &0.10 &0.18 &0.35 &0.52 \\
Materialistic &0.22 &0.36 &0.17 &0.29 &0.10 & 0.17 &0.63 &0.77 &0.19 &0.32 &0.18 &0.31 &0.12 &0.21 &0.30 &0.46 \\
Materialistic MV &0.42 &0.36 &0.23 &0.37 &0.08 & 0.15 &0.64 &0.78 &0.17 &0.29 &0.19 &0.13 &0.14 &0.22 &0.32 &0.29 \\
\bottomrule
\end{tabular}
    }
    \caption{
        Per-scene metrics on the NeRF datasets for the different scenes (columns) and methods (rows). Higher is better.
    }
    \label{tab:qualitative_suppl_nerf}
\end{table*}

\begin{table}[!htp]
    \centering
    \resizebox{\linewidth}{!}{%
        \begin{tabular}{lrrrrrrrrrrr}\toprule
&\multicolumn{2}{c}{\textsc{Garden}} &\multicolumn{2}{c}{\textsc{Kitchen}} &\multicolumn{2}{c}{\textsc{Counter}} &\multicolumn{2}{c}{\textsc{Treehill}} &\multicolumn{2}{c}{\textsc{Bicycle}} \\\cmidrule{2-11}
&mIoU &F1 &mIoU &F1 &mIoU &F1 &mIoU &F1 &mIoU &F1 \\\cmidrule{2-11}
Ours &0.85 &0.92 &0.85 &0.92 &0.74 &0.85 &0.30 &0.46 &0.27 &0.43 \\
SAM2 &0.70 &0.82 &0.62 &0.76 &0.65 &0.79 &0.34 &0.50 &0.22 &0.36 \\
Materialistic &0.34 &0.49 &0.65 &0.79 &0.27 &0.43 &0.16 &0.28 &0.13 &0.23 \\
Materialistic MV &0.13 &0.28 &0.75 &0.86 &0.34 &0.56 &0.25 &0.37 &0.15 &0.25 \\
\bottomrule
\end{tabular}
    }
    \caption{
        Per-scene metrics on our hand-annotated images from the MIPNeRF360 dataset for the different scenes (columns) and methods (rows). For both metrics, higher is better.
    }
    \label{tab:qualitative_suppl_mipnerf}
\end{table}

\mycfigure{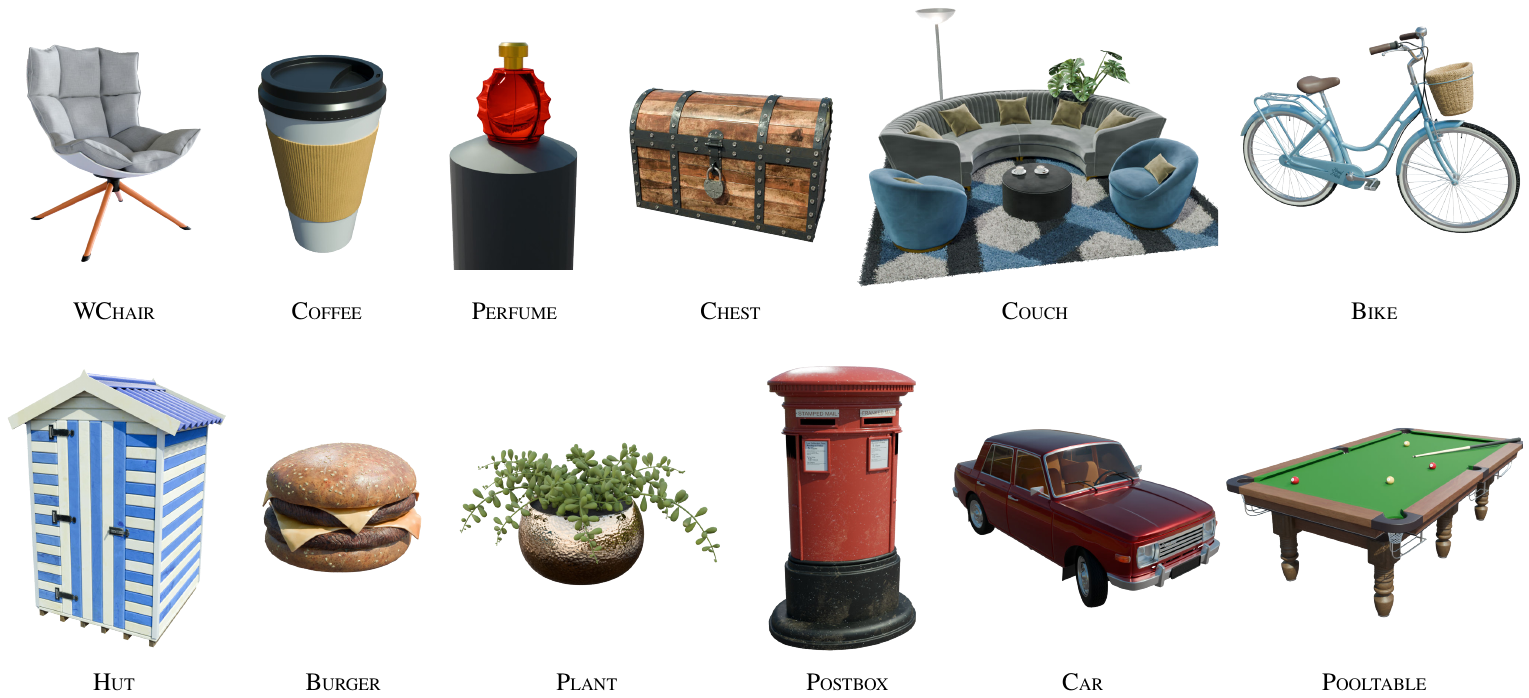}{Our dataset of synthetic objects. Each object has dense material annotations.}



{
    \small
    \bibliographystyle{ieeenat_fullname}
    \bibliography{main}
}